\documentclass[sigconf,nonacm]{acmart}

\usepackage{booktabs}
\usepackage{tikz}
\usepackage{balance}
\usetikzlibrary{arrows.meta,positioning,fit,calc}

\copyrightyear{2026}
\acmYear{2026}
\setcopyright{acmlicensed}
\acmConference[ICAIF '26]{7th ACM International Conference on AI in
Finance}{November 2026}{Milan, Italy}
\acmBooktitle{7th ACM International Conference on AI in Finance (ICAIF '26)}
\acmISBN{978-x-xxxx-xxxx-x/26/11}
\acmDOI{10.1145/nnnnnnn.nnnnnnn}

\newcommand{\thetavec}{\boldsymbol{\theta}}
\newcommand{\Fone}{\textsf{F1}}
\newcommand{\Ftwo}{\textsf{F2}}
\newcommand{\Fthree}{\textsf{F3}}
\newcommand{\Ffour}{\textsf{F4}}

\begin{document}

\title{What Do Foundation Models for Market Microstructure Learn?\\
A Controlled Study with Synthetic Limit Order Books}

\author{Fernando Acero}
\affiliation{%
  \institution{J.P. Morgan QTR AI Research}
  \country{United Kingdom}}
\renewcommand{\shortauthors}{Acero}

\begin{abstract}
Autoregressive transformers trained on limit order book (LOB) message
streams are increasingly proposed as foundation models for market
microstructure. Because the generative process of real markets is
unknown, these models can be judged only by the realism of their output,
leaving open what they actually learn. We make the question answerable by
inverting the setup: we train GPT-style models on order flow from
simulators whose latent parameters are known, so that the Bayes-optimal
predictor, the generating state, and the exact data entropy serve as
references. Measured against them, next-token pretraining on order flow
implements amortized Bayesian inference over market regimes. The learned
predictive matches the exact Bayes optimum to within a millinat per token
and approaches the oracle floor even on a nonstationary family whose
latent state drifts within an episode, a regime that requires filtering.
The latent regime is linearly decodable from frozen activations, well
above baselines, so a frozen-model read-out calibrates simulator
parameters more accurately than windowed maximum likelihood at a fraction
of the cost. In our experiments these capabilities are not produced by
scale: capacity and token volume saturate early, and it is the diversity
of training regimes, not the token count, that determines generalization to
unseen markets. On real NASDAQ data, zero-shot transfer is set by how
well the simulator prior covers real flow: an enriched mixture nearly
closes the gap to an in-domain baseline, and pretraining beats
from-scratch training at every budget across held-out tickers. Known ground truth turns average next-token loss, which is uninformative
near the entropy floor, into an exact comparison against the
Bayes-optimal and oracle predictors.
\end{abstract}



\maketitle

\section{Introduction}
\label{sec:intro}

Sequence models developed for natural language are increasingly applied
to the order flow of the limit order book (LOB). TradeFM trains a 524M
parameter decoder-only transformer on billions of trade events across
more than 9{,}000 equities~\cite{tradefm2026}, and MarS builds a market
simulation engine around a generative order-flow model~\cite{mars2025}.
Since the mechanism that drives a real market is never observed, these
systems can be assessed only through the realism of what they emit:
reproduction of stylized facts and distributional agreement with
held-out data, as systematized by LOB-Bench~\cite{lobbench2025}. Realism
evaluation, however, cannot answer three basic questions about what these
models actually are:

\begin{itemize}
\item[\textbf{RQ1}] How close is the learned next-message distribution to
the best distribution achievable from the observed history?
\item[\textbf{RQ2}] Does the model internally represent the state of the
process that generated the data, or only surface statistics of it?
\item[\textbf{RQ3}] Which resources determine whether these capabilities
emerge: model size, data volume, or properties of the data distribution?
\end{itemize}

\begin{figure}[t]
\centering
\begin{tikzpicture}[
  font=\scriptsize,
  box/.style={draw=black!55, rounded corners=1.5pt, align=center,
              inner sep=2.1pt, minimum height=4.6mm},
  fam/.style={box, fill=blue!9},
  eng/.style={box, fill=gray!14},
  tokbox/.style={box, fill=gray!7},
  mod/.style={box, fill=orange!16},
  ref/.style={box, fill=green!11, text width=33mm},
  arr/.style={-{Stealth[length=1.7mm]}, thick, black!70},
  lbl/.style={font=\tiny\itshape, text=black!55}
]
\node[fam, text width=37mm] (theta)
  {latent parameters $\thetavec \sim p(\thetavec)$\\[-0.5pt]
   {\tiny \Fone{} Poisson $\cdot$ \Ftwo{} Hawkes $\cdot$
    \Fthree{} agents $\cdot$ \Ffour{} nonstat.}};
\node[eng, below=2mm of theta, text width=37mm] (engine)
  {matching engine\\[-1.5pt]
   {\tiny limit order book, price-time priority}};
\node[tokbox, below=2mm of engine, text width=37mm] (tok)
  {message stream $\to$ tokens\\[-0.5pt]
   {\tiny $\underbrace{\textsc{type}\;\textsc{price}\;\textsc{size}\;
     \Delta t}_{\text{1 message}=4\text{ tokens}}$ (512/window)}};
\node[mod, below=2mm of tok, text width=37mm] (gpt)
  {decoder-only transformer\\[-1.5pt]
   {\tiny next-token objective; 25k--32M params}};
\node[ref, right=6mm of theta.north east, anchor=north west] (q1)
  {\textbf{RQ1} behavior\\[-1.5pt]
   {\tiny model predictive $\overset{?}{=}$ exact Bayes
    $p(x_t|x_{<t})$ / oracle $p(x_t|x_{<t},\thetavec)$}};
\node[ref, below=2mm of q1] (q2)
  {\textbf{RQ2} representation\\[-1.5pt]
   {\tiny linear probe: hidden state $\to \thetavec$, vs.\ baselines}};
\node[ref, below=2mm of q2, text width={\dimexpr33mm+4.2pt\relax}, inner xsep=0pt] (q3)
  {\textbf{RQ3} resources\\[-1.5pt]
   {\tiny scaling fit $E \overset{?}{=}$ known entropy floor;\\[-1pt]
    regime diversity $K$}};
\draw[arr] (theta) -- node[lbl, right=1pt] {simulate} (engine);
\draw[arr] (engine) -- node[lbl, right=1pt] {tokenize} (tok);
\draw[arr] (tok) -- node[lbl, right=1pt] {train} (gpt);
\draw[arr] (gpt.east) -- ++(2mm,0) |- (q1.west);
\draw[arr] (gpt.east) -- ++(2mm,0) |- (q2.west);
\draw[arr] (gpt.east) -- ++(2mm,0) |- (q3.west);
\end{tikzpicture}
\caption{Study design. Simulators with known $\thetavec$ generate
message streams through a matching engine. GPT-style models train on the
tokens. Each question has an exact reference: the predictive vs.\ exact
Bayes or oracle (RQ1), probes vs.\ the true $\thetavec$ (RQ2), and
scaling fits vs.\ the true entropy floor (RQ3).}
\label{fig:schematic}
\end{figure}
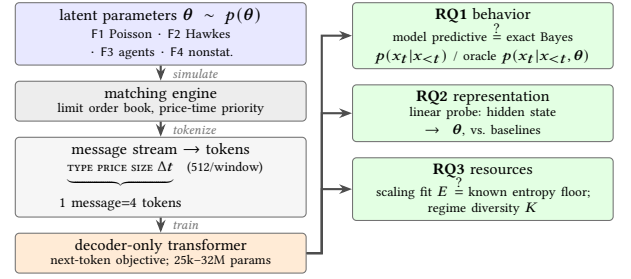

We answer these questions by reversing the usual procedure. Instead of
training on real data and evaluating realism, we train GPT-style models on
message streams produced by market simulators whose latent parameters
$\thetavec$ are known, so that every question above has an exact reference
answer: the Bayes-optimal predictive distribution for RQ1, the true
$\thetavec$ for RQ2, and the exact entropy of the data for RQ3. The
simulators produce complete message-by-message order flow (limit orders,
market orders, cancellations, and inter-arrival times) through a single
matching engine with price-time priority, and span four families of
increasing complexity: independent Poisson flow (\Fone), self-exciting
Hawkes flow (\Ftwo), a heterogeneous agent-based market in which agents
react to the book (\Fthree), and a nonstationary heavy-tailed family
(\Ffour). Figure~\ref{fig:schematic} summarizes the design. All models
are trained on tokenized messages with a standard next-token objective,
exactly as the large-scale systems above are trained on real data.

The controlled setting yields the following answers.

\emph{Answer to RQ1: the trained model is indistinguishable from the
Bayes-optimal forecaster.} On \Fone{}, where the Bayes-optimal next-token
distribution can be computed exactly, the trained model matches it to
$3.4\times10^{-4}$ nats (natural-log units of information) per token at
512 messages of context
(bootstrap 95\% confidence interval $[-7.3,+0.2]\times10^{-4}$). On the
three families without closed-form structure the model converges to exact
oracle floors, including \Ffour{}, where the latent state drifts within an
episode, so matching the oracle demands online filtering, not inference
over a fixed regime. This extends the in-context-learning-as-Bayes
result~\cite{xie2021explanation,muller2022transformers}, shown so far for fixed latent parameters, to online filtering, and
verifies it against an exact oracle at millinat precision rather than by
asymptotic argument.

\emph{Answer to RQ2: the latent regime is encoded in the hidden states.}
Linear regressions from frozen hidden states
(probes~\cite{alain2016understanding}) recover the
agent-mix composition of \Fthree{} with $R^2$ of 0.92--0.97, reliably
above regressions on 22 hand-crafted summary statistics of the same
window (0.52--0.92) and probes on an untrained network of the same
architecture (0.42--0.81). The advantage is largest at short contexts,
where the defining statistics have not yet accumulated. Parameters that
are recoverable from simple statistics are recovered equally well by all methods. The trained representation has an advantage only on the
dynamical parameters. A practical consequence is calibration at
negligible cost: a linear read-out of the frozen model recovers Hawkes
parameters with 2.7--4.5$\times$ lower error than windowed maximum
likelihood and also beats a supervised regressor trained specifically for
the task.

\emph{Answer to RQ3: capacity and data volume saturate early; regime diversity is the limiting resource.} Fitting Chinchilla-style scaling
laws~\cite{hoffmann2022chinchilla} to a grid of model sizes and token
budgets recovers an irreducible
constant within 2 millinats of the known floor, and, in these simulator
families, the floor is reached by models with roughly $10^5$ parameters. In contrast, holding tokens
fixed and varying the number of distinct market regimes $K$ in the
training data changes generalization qualitatively: a model trained on
one regime fails on fresh regimes (3.57 nats versus a 1.34 floor), and
performance converges to the unlimited-diversity limit only near
$K{=}256$. Since the entropy floor of real markets is unknown, these
results caution against reading capability growth into scaling exponents
fitted to real LOB data, and identify regime coverage rather than token
count as the quantity to maximize.

Finally, we test what carries over to real markets. Generated flow
maintains valid books over thousands of closed-loop steps. A single mixture-trained model matches every specialist within millinats. On one day of NASDAQ data from the LOBSTER reconstruction of the
ITCH feed~\cite{huang2011lobster}, zero-shot transfer depends on prior coverage: adding the heavy-tailed nonstationary family \Ffour{} nearly
closes the size gap to real flow, and yields an initialization that beats
from-scratch training at every budget across three held-out tickers.

To the best of our knowledge, this is the first study to characterize what
autoregressive foundation models for market microstructure learn against
known simulator ground truth. Our contributions are: (i) a controlled
experimental framework that
couples message-level LOB simulators having known ground truth with a
foundation-model training pipeline, including exact Bayes predictives and
oracle floors; (ii) exact verification that next-token pretraining on
order flow implements amortized Bayesian inference and filtering in a
market-microstructure setting; (iii) a probing study of what is encoded
where, validated against falsification baselines, with parameter
calibration as an application; (iv) controlled scaling and diversity
results that reinterpret what scaling laws measure when the entropy floor
is known; and (v) a per-field real-data transfer analysis showing that zero-shot transfer depends on the simulator prior's coverage of
distributional properties of real flow.

\section{Related Work}
\label{sec:related}

\textbf{Generative models of order flow.} Deep generative models of LOB
message streams have progressed from single-asset token-level models to
large cross-asset systems. TradeFM trains a 524M parameter decoder-only
transformer on more than $10^{10}$ tokens of trade events from over 9{,}000
US equities, using scale-invariant features and a unified tokenization, and
reports 2--3$\times$ lower distributional error than compound Hawkes
baselines together with zero-shot generalization to out-of-distribution
markets~\cite{tradefm2026}. MarS couples an order-level generative model to
a simulation engine for downstream what-if analysis~\cite{mars2025}.
Token-level autoregressive models of message flow were introduced at
smaller scale by \citet{nagy2023generative}.
Evaluation in this literature is distributional: LOB-Bench formalizes
realism scoring of generated message streams against LOBSTER
data~\cite{lobbench2025,huang2011lobster}, and earlier work argued for
universal, transferable structure in price formation across
assets~\cite{sirignano2021universal}. All of these evaluations compare
model output to data whose generative process is unknown. Our study is complementary: by making the generator known, it converts
questions about what such models learn into questions with exact
answers.

\textbf{In-context learning as Bayesian inference.} The phenomenon we
verify is anticipated by a substantial literature outside finance.
\citet{xie2021explanation} explain in-context learning as implicit
Bayesian inference over a latent concept variable. Prior-data fitted
networks train transformers on samples from a prior precisely so that they
approximate the posterior predictive~\cite{muller2022transformers}, with
TabPFN as a practical instantiation~\cite{hollmann2023tabpfn}. Earlier work established that meta-trained sequential agents converge to
Bayes-optimal behavior~\cite{ortega2019meta,mikulik2020meta}. A related
line studies which algorithms transformers implement in context for
function classes~\cite{garg2022what,akyurek2023what,vonoswald2023transformers},
and \citet{raventos2023pretraining} show that pretraining task diversity
gates the emergence of in-context learning for regression. Relative to
this literature our contribution is not the phenomenon but its exact,
domain-relevant verification: a market setting with an engine in the loop,
structured multi-field tokens, closed-form conjugate floors at millinat
precision rather than asymptotic arguments, an extension to filtering of a
time-varying latent state, and consequences specific to finance
(simulator calibration and diagnosis).

\textbf{Classical microstructure models and simulation.} Our simulator
families are built from standard ingredients: zero-intelligence order
flow in a continuous double auction~\cite{smith2003statistical},
mutually exciting Hawkes processes, the standard model for event-level
microstructure econometrics~\cite{bacry2015hawkes}, order-flow imbalance
as a price predictor~\cite{cont2014price}, and simulation by Ogata
thinning~\cite{ogata1981lewis}. Engine-based multi-agent market
simulators such as ABIDES~\cite{byrd2020abides} pursue realism for downstream applications. Our engine is intentionally minimal so that
exact references remain computable. Our probing methodology follows the
linear-probe literature in interpretability~\cite{alain2016understanding,
belinkov2022probing}. Deep learning on LOB data for prediction
is well established~\cite{zhang2019deeplob,sirignano2021universal}.
Scaling analyses follow the parametric form of
\citet{kaplan2020scaling,hoffmann2022chinchilla}. Our contribution there is the comparison of the fitted irreducible constant against a known
floor, which is unavailable in any natural-data setting.
\vspace{-1mm}

\section{Experimental Setup}
\label{sec:setup}

We describe the shared engine and tokens (Section~\ref{sec:engine}),
the simulator families (Section~\ref{sec:families}), and the models and
training (Section~\ref{sec:training}).

\subsection{Market engine, messages, and tokens}
\label{sec:engine}

All simulator families share one matching engine, similar to multi-agent market simulators such as ABIDES~\cite{byrd2020abides} but kept to the minimum needed for controlled analysis. The engine
maintains a single-asset limit order book on an integer price grid
(ticks): each price level holds a first-in-first-out queue of resting
orders, and execution follows price-time priority. A limit order priced
at or through the best quote on the opposite side executes immediately
for the crossing portion and rests otherwise, so the book can never
cross; market orders execute against resting orders and never walk past
available liquidity; cancellations remove a uniformly random resting
order from the chosen side. The engine is deterministic given the
message sequence and the cancellation draws, which is what allows the
oracle computations described below.

Every event in a simulation is a \emph{message} with four fields:
an event type, a price, a size, and the inter-arrival time
$\Delta t$ since the previous message. There are six event types: buy and
sell limit orders, buy and sell market orders, and cancellations on the
bid and on the ask side. Prices of limit orders are recorded as the
offset $d\ge 0$, in ticks, of the order relative to the best quote on the
opposite side of the book. This makes the representation independent of
the absolute price level. An episode of a family first draws its latent
parameter vector $\thetavec$ from the family's prior
(Table~\ref{tab:dgp}), seeds the book, runs 2{,}000 unrecorded burn-in
messages, and then records $16{,}384$ messages.

Messages are tokenized as four consecutive tokens in fixed order
[\textsc{type}, \textsc{price}, \textsc{size}, $\Delta t$]: 6 type
tokens; 22 price tokens (offsets $d=0,\dots,19$, one token for
$d\ge 20$, and one structural not-applicable token used by market orders
and cancellations); 8 size tokens with bin edges
$\{1,2,3,5,9,17,33,65\}$; and 16 $\Delta t$ tokens log-spaced over
$[10^{-4},30]$ seconds with the end bins absorbing values outside the
range. With a beginning-of-sequence token the vocabulary has 53 symbols.
Training sequences are non-overlapping windows of $W=512$ consecutive
messages (2{,}049 tokens). Bin edges are identical across families, and
no bin dominates any field (the largest $\Delta t$ bin carries 15.9\% of
its field's mass). Losses throughout are reported in nats per token.

\subsection{Simulator families}
\label{sec:families}

The four families form a ladder of increasing complexity: independent
flow, self-excitation, agent feedback through the book, and
nonstationarity with heavy tails.

\textbf{\Fone{} (Poisson).} Conditional on $\thetavec$, every field of
every message is drawn independently. The event type is categorical with
probability vector $\pi$ over the six types; the limit-order offset is
geometric, $d\sim\text{Geom}(p_{\mathrm{geo}})$; sizes are
$1+\text{Pois}(\bar{s})$ with mean parameter $\bar{s}$ (market orders use
mean $\bar{s}/8$); and inter-arrival times are exponential with rate
$\Lambda$, $\Delta t\sim\text{Exp}(\Lambda)$. The prior on $\pi$ is
hierarchical: the total limit-order probability
$g\sim\text{Beta}(137,103)$ is split between buy and sell by a
$\text{Dir}(4,4)$ draw, and the remaining mass is split among
market orders and cancellations by a $\text{Dir}(2,2,3,3)$ draw. This
construction keeps the posterior over $\pi$ in closed form and, together
with two structural rules described below, produces stable books over
essentially the entire prior. As a result the exact Bayes-optimal
next-token distribution, that is, the predictive obtained by integrating
over the posterior of $\thetavec$ given the observed tokens, is
computable: the type field by closed-form posterior updates, and the
remaining scalar parameters ($\Lambda$, $p_{\mathrm{geo}}$, $\bar{s}$) by
512-point numerical integration over their priors.

\textbf{\Ftwo{} (Hawkes).} Event types follow a six-dimensional Hawkes
process, a point process in which each event raises the arrival intensity of future events. The excitation decays exponentially at rate $\beta$ and
its overall strength is the branching ratio $a$, the expected number of
direct descendants of one event. The cross-excitation pattern between the
six types is a fixed matrix, and $a$ equals the spectral radius of the
branching matrix, so $a\sim U(0.1,0.85)$ guarantees stability. Simulation
uses the exact thinning algorithm of \citet{ogata1981lewis}, verified by
Kolmogorov--Smirnov tests on time-rescaled residuals. Offsets and sizes
follow the \Fone{} laws. For this family the exact Bayes predictive is
not available in closed form, so we evaluate against the \emph{oracle}:
the conditional distribution of the next token given the history
\emph{and} the true $\thetavec$, computed by replaying the intensity
recursion. The oracle is the entropy floor of the family, which no predictor
can beat on average.

\textbf{\Fthree{} (agent-based).} Messages are emitted by a population of
four agent types: noise traders (\Fone{}-like flow), momentum traders
that trade in the direction of the mid-price change over the last 50
messages, mean reverters that trade against it, and a market maker that
keeps 5--12 resting orders, at least five per side. The mix is
$w\sim\text{Dir}(1,1,1,1)$, and agents observe and react to the book, so
the flow feeds back on itself. Only the public message stream is tokenized. Agent identities are hidden. The oracle is the exact
next-message distribution given the full simulator state, computed
analytically (no Monte Carlo) by summing the four agent branches over
their next actions.

\textbf{\Ffour{} (nonstationary, heavy-tailed).} \Ftwo{} dynamics with
three additions: sizes concentrate on a single lot with probability
$p_{\mathrm{lot}}$ and otherwise follow a Pareto tail with index
$\alpha_s$; the baseline intensity is modulated by a lognormal factor
with volatility $\sigma_{\mathrm{mod}}$, resampled every 800 messages;
and rare intensity jumps decay geometrically. The latent state therefore
moves within an episode, and the oracle (again exact, integrating the
modulated Hawkes intensity over the inter-arrival grid) is a filtering
reference: matching it requires tracking the current state, not just
identifying fixed parameters.

\begin{table}\centering\small
\caption{Simulator families (latent parameters and priors) and model
sizes. All families
share the engine, tokenizer, and episode structure (2{,}000 burn-in +
16{,}384 recorded messages). LogU = log-uniform; Dir = Dirichlet.}
\label{tab:dgp}\label{tab:models}
\begin{tabular}{llll}\toprule
family & parameter & prior & role \\ \midrule
all & $\Lambda$ (rate, s$^{-1}$) & LogU$(0.5,50)$ & activity \\
\Fone,\Ftwo & $g$ (LO mass) & Beta$(137,103)$ & book stability \\
 & bid/ask split & Dir$(4,4)$ & side asymmetry \\
 & MO/CX split & Dir$(2,2,3,3)$ & flow mix \\
 & $p_{\mathrm{geo}}$ (offsets) & $U(0.15,0.6)$ & book depth \\
 & $\bar{s}$ (sizes) & LogU$(1,60)$ & order sizes \\
\Ftwo,\Ffour & $a$ (branching) & $U(0.1,0.85)$ & self-excitation \\
 & $\beta$ (decay, s$^{-1}$) & LogU$(0.5,20)$ & memory length \\
\Fthree & $w$ (agent mix) & Dir$(1,1,1,1)$ & composition \\
\Ffour & $p_{\mathrm{lot}}$ & $U(0.3,0.9)$ & lot concentration \\
 & $\alpha_s$ (size tail) & $U(1.2,2.5)$ & heavy tails \\
 & $\sigma_{\mathrm{mod}}$ & $U(0.3,0.9)$ & nonstationarity \\ \midrule
\multicolumn{4}{l}{OOD priors: \Ftwo{} $a\sim U(0.85,0.95)$; \Fthree{}
$w\sim$Dir$(1,8,1,1)$ (momentum corner).}\\
\bottomrule
\end{tabular}

\medskip

\begin{tabular}{lcccc}\toprule
\multicolumn{5}{l}{\emph{Model sizes (non-embedding parameters)}}\\
 & $d_{model}$ & layers & heads & params \\ \midrule
m0 / m1 / m2 & 32 / 64 / 96 & 2 / 3 / 3 & 2 / 4 / 4 & 25k / 150k / 336k \\
S0 / S1 & 128 / 256 & 4 / 6 & 4 / 8 & 0.79M / 4.7M \\
S2 / S3 & 384 / 512 & 8 / 10 & 8 / 8 & 14.2M / 31.5M \\
\bottomrule
\end{tabular}
\end{table}

\begin{figure*}[t]
\centering
\includegraphics[width=\textwidth]{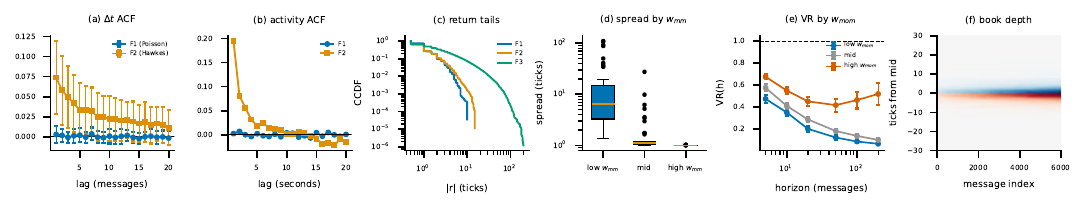}
\caption{Statistical properties of the families (40 episodes each; 150
for \Fthree{}). (a) Inter-arrival and (b) one-second activity
autocorrelation: \Ftwo{} clusters, \Fone{} does not. (c) \Fthree{} has
heavy return tails. (d) Market-maker weight tightens spreads (1.0 vs.\
17.4 ticks across terciles). (e) Momentum weight raises variance ratios
(trend persistence; 1 = random walk). (f) Book depth in one episode.}
\label{fig:stylized}
\vspace*{-2mm}
\end{figure*}

\textbf{Structural rules and disclosures.} Two rules keep books stable
for asymmetric parameter draws: cancellations become non-tokenized no-ops
when a side holds ten or fewer orders, and market-order sizes are smaller
than limit-order sizes in \Fone{} and \Ftwo{}. The no-op rule is the single deviation from exact
conditional independence in \Fone{}. It affects 0.7--1.4\% of messages,
and Section~\ref{sec:bayes} bounds its effect on the Bayes comparison
below a millinat. Episodes in which a book side empties more than 0.1\%
of the time are discarded and redrawn (0--2\% of draws; 17\% for
\Ffour{}). The behavioral constants of \Fthree{} were chosen once, so that
the families show measurably distinct dynamics
(Figure~\ref{fig:stylized}), and then held fixed: momentum and
mean-reversion agents send a market order with probability 0.4 and
otherwise a limit order, one tick inside the best quote or at a
Geom(0.4) offset, respectively; the market maker cancel-replaces with
probability 0.05; and noise traders use Geom(0.1) offsets and mean
size 8, with market orders twice as large.

\subsection{Models, training, and evaluation splits}
\label{sec:training}

Models are GPT-2-style decoder-only transformers (pre-layer-norm, learned
positional embeddings, weight-tied output head) in seven sizes, denoted
m0--m2 and S0--S3 (Table~\ref{tab:dgp}, bottom), spanning 25k to 31.5M
non-embedding parameters. The main models (size S2, 14.2M parameters)
train for a single pass over 1.6B tokens per family with AdamW
($\beta=(0.9,0.95)$, weight decay $0.1$ on matrices), a peak learning
rate of $6\times10^{-4}$ (sizes m0--S1) or $3\times10^{-4}$ (S2--S3) with
a 300-step warmup and cosine decay to 10\% of peak, gradient clipping at
$1.0$, a batch of 64 sequences, and bf16 precision, at 131k tokens per
second on a single-GPU workstation. The complete study comprises roughly
100 runs and 55 GPU-hours. Training, in-distribution validation, out-of-distribution
validation, and probing sets use disjoint pools of $\thetavec$. The out-of-distribution priors shift the excitation range (\Ftwo{}) or
concentrate the agent mix on a momentum-heavy corner (\Fthree{},
Table~\ref{tab:dgp}). Seed robustness was checked by repeating the main
runs with a second seed, and the largest seed-to-seed difference in any
reported validation loss is 1.6 millinats.
\vspace{-1mm}

\section{Results}
\label{sec:results}

We present the results in the order of the research questions: the
behavior of the learned predictive distribution against exact references
(Section~\ref{sec:bayes}), the internal representation that supports it
(Section~\ref{sec:probes}), the training resources that produce it
(Section~\ref{sec:scaling}), and the extent to which the findings extend
to generation and to real data (Section~\ref{sec:transfer}).

\subsection{RQ1: The learned predictive distribution matches the Bayes optimum}
\label{sec:bayes}

\begin{figure*}[t]
\centering
\scalebox{1}[0.92]{\includegraphics[width=\textwidth]{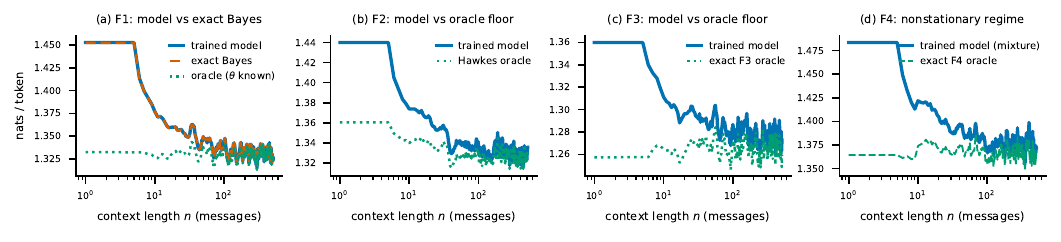}}
\caption{Mean prediction loss vs.\ window position $n$ for S2 models
(400 held-out episodes per family; 9-point smoothing). Solid: trained
model; dashed: exact Bayes (\Fone{}); dotted: oracle given $\thetavec$.
The model matches exact Bayes on \Fone{} (a) and reaches the oracle
floors elsewhere, including the moving latent state of \Ffour{} (d),
which requires filtering. The model in (d) is the four-family mixture.}
\label{fig:incontext}
\vspace*{-2mm}
\end{figure*}

We first ask how close the trained model's next-token distribution is to
the best achievable one. Figure~\ref{fig:incontext} shows, for each
family, the mean prediction loss at each position $n$ within a
512-message window, together with the exact reference for that family.

On \Fone{} the reference is the exact sequential Bayes predictive,
\begin{equation}
\label{eq:bayes}
p(x_{n+1}\mid x_{1:n})=\int_{\Theta} p(x_{n+1}\mid x_{1:n},\thetavec)\,
p(\thetavec\mid x_{1:n})\,d\thetavec,
\end{equation}
obtained by integrating the next-token law over the posterior
$p(\thetavec\mid x_{1:n})\propto p(\thetavec)\prod_{i=1}^{n}
p(x_i\mid x_{1:i-1},\thetavec)$, which is updated after every observed
token. This is the best any predictor without knowledge of $\thetavec$
can do on average. For \Ftwo{}--\Ffour{} the integral in
Eq.~\eqref{eq:bayes} is intractable, so we compare instead against the
oracle $p(x_{n+1}\mid x_{1:n},\thetavec^\star)$ at the true
$\thetavec^\star$ (Section~\ref{sec:families}), the entropy floor no
predictor can beat. The trained model's loss curve coincides
with the Bayes curve over the entire context range. The gap at $n=512$ is
$-3.4\times10^{-4}$ nats per token (episode-bootstrap 95\% confidence
interval $[-7.3,+0.2]\times10^{-4}$), and mean validation loss is within
0.11\% of the Bayes bound. The gap is statistically indistinguishable
from zero or slightly negative. A negative gap is consistent with
the one modeling effect that would produce it, namely that the
closed-form predictive ignores the cancellation floor rule of
Section~\ref{sec:families} while the model learns its consequences from
data. Either way, the comparison bounds the effect of that rule below a
millinat.

The same convergence holds where no closed form exists: gaps to the exact
oracles at $n{=}512$ are $0.0026$ nats (\Ftwo{}), $0.0110$ (\Fthree{}),
and $0.0085$ (\Ffour{}). Unlike the \Fone{} Bayes reference, these
oracles know the true state, so their gaps, an order larger than
\Fone{}'s, are the unavoidable cost of inferring it, not model
suboptimality. The \Ffour{}
case extends the claim beyond fixed parameters: because the latent
intensity state changes within an episode,
staying near the oracle requires continuously re-estimating the current
state from recent messages, and the model does so within 9 millinats.

Table~\ref{tab:ladder} places these results against classical sequence
models on the same data. A backoff 5-gram over the token stream captures
short-range structure but sits 0.17--0.24 nats above the transformer, which reaches the floor. The gap reflects inference over the latent
regime, which needs hundreds of messages of context that a short-context
model lacks.

\begin{table}\centering\small
\caption{Validation loss (nats per token) of a backoff 5-gram model over
the token stream, the trained transformer (size S2,
Table~\ref{tab:models}), and the exactly known irreducible floor of each
family. The 5-gram captures short-range structure, and its gap to the
transformer measures the value of long-context inference over the latent
regime. \Ffour{} is omitted: its nonstationary floor is not a single
per-token constant.}
\label{tab:ladder}
\begin{tabular}{lccc}\toprule
family & 5-gram & transformer (S2) & irreducible floor \\ \midrule
F1 & 1.575 & 1.3379 & 1.3365 (exact Bayes-window) \\
F2 & 1.568 & 1.3359 & 1.3271 (oracle) \\
F3 & 1.439 & 1.2718 & 1.2552 (oracle) \\
\bottomrule\end{tabular}\vspace*{-2mm}
\end{table}

\subsection{RQ2: The latent regime is encoded in the hidden states}
\label{sec:probes}

The behavioral result of Section~\ref{sec:bayes} implies that the model
must carry information about the regime, and we now locate it. On its own, decodability shows only that the regime information is \emph{present}, not that the model computes with it. That link comes from the Bayes-optimal behavior of Section~\ref{sec:bayes}, which is unattainable without using the regime, and a fully causal account via representational intervention remains future work. We fit ridge
regressions (linear probes~\cite{alain2016understanding,belinkov2022probing})
from frozen residual-stream activations to each component of $\thetavec$,
at every layer and at context lengths $m\in\{8,32,128,512\}$ messages,
with train and test sets split at the episode level. Two falsification baselines control for trivial
explanations: identical probes on an untrained network of the same
architecture (testing whether random features suffice), and ridge
regression on 22 hand-crafted summary statistics of the same $m$ messages
(testing whether the information is available in surface statistics). For
\Ftwo{} we add the classical estimator, windowed maximum likelihood.

\begin{table}\centering\small
\caption{Test $R^2$ of linear probes from the trained S2 model's
activations to latent generator parameters at $m{=}512$ messages of
context, against falsification baselines. Dynamical parameters:
mean $\pm$ sd over 5 episode-grouped CV folds at the best layer. Static
parameters (below the rule) are recovered by all methods and shown from a
single split. MLE = windowed Hawkes maximum likelihood ($R^2$; negative =
worse than predicting the mean).}
\label{tab:probe}
\begin{tabular}{lcccc}\toprule
target & trained & random-init & summary stats & MLE \\ \midrule
$a$ (\Ftwo) & $\mathbf{0.83}\pm.02$ & $0.66\pm.02$ & $0.73\pm.02$ & $-0.18$ \\
$\log\beta$ (\Ftwo) & $\mathbf{0.70}\pm.03$ & $0.07\pm.04$ & $0.42\pm.03$ & $-4.89$ \\
$w_{mom}$ (\Fthree) & $\mathbf{0.92}\pm.01$ & $0.42\pm.05$ & $0.52\pm.03$ & -- \\
$w_{mm}$ (\Fthree) & $\mathbf{0.97}\pm.01$ & $0.81\pm.06$ & $0.92\pm.01$ & -- \\ \midrule
$\log\Lambda$ (\Ftwo) & 0.98 & 0.93 & 0.97 & -- \\
$p_{geo}$ (\Ftwo) & 0.95 & 0.95 & 0.87 & -- \\
\bottomrule
\end{tabular}
\vspace*{-3mm}
\end{table}

\begin{figure*}[t]
\centering
\includegraphics[width=\textwidth]{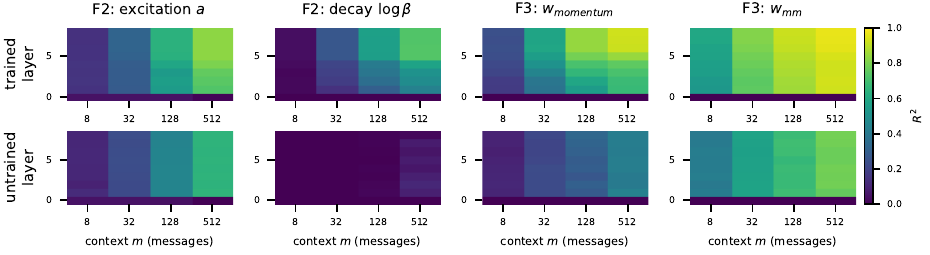}
\caption{Probe $R^2$ by layer (0 = embeddings) and context length $m$,
for the trained S2 model (top) and an untrained network (bottom); yellow
= accurate recovery. Composition weights peak at the deepest layer, and the advantage of training is largest at short contexts.}
\label{fig:probes}
\vspace*{-2mm}
\end{figure*}

Three regularities emerge (Table~\ref{tab:probe},
Figure~\ref{fig:probes}). First, the dynamical parameters separate the
trained model from every baseline by several fold-level standard
deviations: momentum weight $R^2=0.92\pm.01$ against $0.52\pm.03$
(summary statistics) and $0.42\pm.05$ (untrained network); excitation $a$
reaches $0.83\pm.02$ against $0.73\pm.02$ and $0.66\pm.02$, and decay
$\log\beta$ reaches $0.70\pm.03$ against $0.42\pm.03$ and $0.07\pm.04$. The market-maker
weight is the narrowest separation ($0.97\pm.01$ vs.\ $0.92\pm.01$)
because its spread signature is easy to read from the window. Second, the
advantage concentrates at short contexts (momentum weight at $m{=}32$:
$0.59$ vs.\ $0.26$): the model extracts regime information faster than
the defining statistics accumulate. Third, parameters that simple
statistics recover (event rate, offset decay) are recovered equally well by the untrained network. Decodability alone is therefore not evidence of learning, and we report those rows as controls. Composition weights peak at the
deepest layer, and probes on intermediate checkpoints show that most
decodability is in place after a quarter of training.

The representation has a direct use. Calibrating a Hawkes or agent-based
simulator to a window is a standard task, usually done by maximum
likelihood or method-of-moments fitting.
Table~\ref{tab:calibration} compares three estimators of
$(a,\log\beta)$ on held-out 512-message windows: a linear read-out of the
frozen pretrained model, windowed maximum likelihood, and a supervised
multilayer perceptron trained on the summary statistics for this exact
task. The frozen read-out is the most accurate and the cheapest. The
maximum-likelihood implementation is sound: on windows eight times
longer it recovers both parameters nearly perfectly ($R^2$ 0.97/0.99).
It fails at 512 messages, which carry little information, whereas the
pretrained model has amortized the prior over regimes into its weights.

\begin{table}\centering\small
\caption{Calibration of the \Ftwo{} parameters $(a,\log\beta)$ from
single 512-message windows: root-mean-square recovery error on held-out
windows and wall-clock time per window (same GPU). FM probe: linear
read-out of the frozen pretrained S2 model. NPE-MLP: a multilayer
perceptron trained on the 22 summary statistics for this task. Hawkes
MLE: windowed maximum likelihood.}
\label{tab:calibration}
\begin{tabular}{lccc}\toprule
 & FM probe & NPE-MLP & Hawkes MLE \\ \midrule
RMSE $a$ & \textbf{0.087} & 0.114 & 0.236 \\
RMSE $\log\beta$ & \textbf{0.556} & 0.727 & 2.490 \\
sec / window & 0.02 & $\sim$0.02 & 5.5 \\
\bottomrule\end{tabular}\vspace*{-2mm}
\end{table}

\subsection{RQ3: Diversity, not capacity or volume, is the limiting
resource}
\label{sec:scaling}

Given that trained models reach the Bayes floor, we ask what it takes to
get there. We train a grid of model sizes and token budgets on \Ftwo{}
(16 points, 25M--1.6B tokens) and \Fone{} (a smaller grid), and fit a
Chinchilla-style scaling law, the saturating power law
$L(N,D)=E+A\,N^{-\alpha_N}+B\,D^{-\alpha_D}$ of
\citet{hoffmann2022chinchilla}, where $N$ is the parameter count, $D$
the number of training tokens, $A,B$ coefficients, and $E$ the fitted
irreducible loss. In language modeling such fits extrapolate capability with scale. Here, by contrast, the fitted $E$ can be compared to the true
irreducible loss of the data.

\begin{figure*}[t]
\centering
\includegraphics[width=\textwidth]{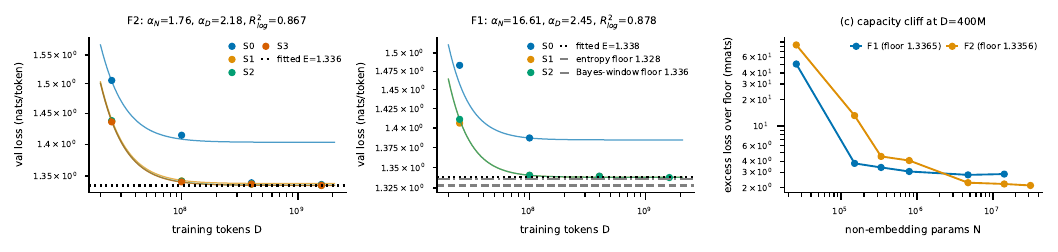}
\caption{Scaling with a known floor. (a) \Ftwo{}, (b) \Fone{}: loss
vs.\ training tokens for the S-series sizes S0--S3 (points) with the
joint Chinchilla-style fit (curves); horizontal lines mark the fitted
$E$ and the known floors. (c) Excess loss over the floor at $D{=}400$M
vs.\ parameter count, now including the micro sizes m0--m2 that expose
the low-capacity regime: capacity binds only below roughly $10^5$
parameters.}
\label{fig:scaling}
\vspace*{-2mm}
\end{figure*}

Two conclusions follow (Figure~\ref{fig:scaling}). First, the fit
recovers the true irreducible loss: for \Fone{}, $E=1.3384$ against an
exactly computable floor of $1.3365$. The floor that matters is not the
conditional entropy given $\thetavec$ (1.3282) but that entropy plus the
unavoidable cost of inferring $\thetavec$ within a 512-message window,
approximately 8 millinats per token, and the fit recovers this larger
number. Second, within these simulator families capacity and data volume
saturate early: a model with $1.5\times10^5$ parameters is within 4
millinats of the floor at 400M tokens, and all models from 0.8M to 31.5M
parameters are statistically indistinguishable at every budget. As a consequence the exponents
$\alpha_N,\alpha_D$ are poorly identified ($R^2_{\log}=0.87$): the loss
surface is a floor plus a fast transient rather than a broad power law.
Because the entropy floor of real markets is unknown, this observation
cautions against interpreting scaling exponents fitted to real LOB data
as measures of capability growth.

\begin{figure}[t]
\centering
\includegraphics[width=\linewidth]{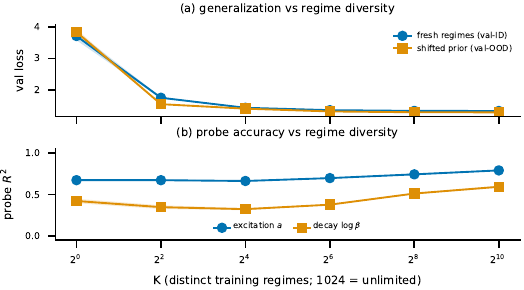}
\caption{Diversity ablation: S1 models, fixed 400M-token budget, $K$
distinct regimes in training (two seeds; bands span seeds). (a) Loss on
unseen regimes and under a shifted prior. (b) Best-layer probe $R^2$ at
$m{=}512$: decodability of the harder parameter is created by
diversity.}
\label{fig:diversity}
\vspace*{-2mm}
\end{figure}

What does not saturate is diversity. We fix the token budget at 400M and
vary the number of distinct market regimes $K$ in the training data by
cycling episodes over $K$ fixed draws of $\thetavec$ (\Ftwo{}, two
seeds). Loss on fresh regimes falls monotonically from 3.57 nats at
$K{=}1$, where the model memorizes its single regime and fails on new
ones, to 1.34 at $K{=}256$, indistinguishable from training with a fresh
$\thetavec$ every episode (Figure~\ref{fig:diversity}a). Probing
separates two sources of decodability (Figure~\ref{fig:diversity}b): for
the excitation $a$, probe accuracy at $K{\le}16$ equals the
untrained-network level, and only diversity beyond that raises it. For the decay $\log\beta$, accuracy climbs from 0.32 to 0.59 with $K$ against
an untrained-network level of 0.11. The capability documented under RQ1
and RQ2 is therefore created by regime diversity, consistent with the
task-diversity threshold that \citet{raventos2023pretraining} report for
in-context regression, here in a setting with a market mechanism in the
loop.

\subsection{External validity: generation, generalists, and real data}
\label{sec:transfer}

The preceding sections establish what the models learn in the controlled
setting. We now test how far these findings extend to generation and to
real data.

\textbf{Closed-loop generation.} Sampling from the model and executing
each sampled message in the engine for 2{,}048 steps, all 100 rollouts
per family maintain a valid two-sided book throughout. Wasserstein-1
distances between generated and held-out episodes are small for
\Fone{}/\Ftwo{} (spread distributions differ by 0.011/0.025 ticks) but
reveal a clear closed-loop gap for \Fthree{} (4.9 ticks), where
feedback between the model and the book compounds small errors. We
report this as a limitation of the learned agent-based model.

\textbf{Generalists match specialists.} A single model trained on the
mixture of families matches each specialist on its home family within
0.7--6 millinats (Table~\ref{tab:ood}), while specialists transfer
asymmetrically (the \Fthree{} specialist scores 3.16 nats on \Fone{}).
At this scale there is no cost to training one general model.

\begin{table}\centering\small
\caption{Cross-family zero-shot val loss (nats/token): every model
evaluated on every family. The mixture generalist matches each specialist
on its home family within millinats.}
\label{tab:ood}
\begin{tabular}{lccc}\toprule
model $\backslash$ eval & F1 & F2 & F3 \\ \midrule
F1 & 1.337 & 1.367 & 2.086 \\
F2 & 1.339 & 1.338 & 2.082 \\
F3 & 3.164 & 3.271 & 1.272 \\
MIX & 1.337 & 1.339 & 1.277 \\
\bottomrule\end{tabular}\vspace*{-2mm}
\end{table}

\textbf{Real data: transfer depends on prior coverage.} We evaluate
mixture models zero-shot on one day of NASDAQ LOBSTER message data
(AMZN, MSFT, INTC, AAPL, GOOG; 2012-06-21)~\cite{huang2011lobster},
mapping each
NASDAQ event to the token schema (submissions to limit orders, deletions
and partial cancellations to cancellations, executions to market orders
with same-timestamp executions merged; offsets in ticks from the opposite
best; sizes in round lots; $\Delta t$ in seconds). This is a demanding
test: the models have never seen a real message.
Table~\ref{tab:transfer} compares the mixture over
\Fone{}--\Fthree{} with the mixture that also includes \Ffour{}, whose
defining properties (lot-concentrated heavy-tailed sizes, intraday
nonstationarity, bursts) are exactly the properties of real flow that the
simpler families lack. The effect is large and field-specific: price
placement transfers well from either mixture, while including \Ffour{}
cuts the size loss almost to the in-domain unigram (1.60 to 0.91 nats,
closing 98\% of the gap) and lowers inter-arrival loss by up to 0.55 nats
per ticker. Fine-tuning the enriched model on an AMZN+MSFT slice then
beats from-scratch training at every budget, averaged over three fully
held-out tickers (INTC, AAPL, GOOG), and the advantage grows as the slice
shrinks (Table~\ref{tab:transfer}, bottom).
Zero-shot transfer is therefore governed by whether the simulator prior
covers the distributional properties of real flow, and pretraining on a
sufficiently rich synthetic prior is a useful initialization when real
data are scarce.

\begin{table}\centering\small
\caption{Real-data transfer (NASDAQ LOBSTER, 2012-06-21; nats/token).
Top: zero-shot per-field loss averaged over five tickers for the
original mixture, the \Ffour{}-enriched mixture, and a unigram fit on the same day. Bold marks the better transfer model per field (the unigram is
an in-domain reference, not a transfer model). Bottom: fine-tuning the
enriched model on an AMZN+MSFT slice vs.\ from-scratch training on it,
averaged over three fully held-out tickers (INTC, AAPL, GOOG).}
\label{tab:transfer}
\begin{tabular}{lcccc}\toprule
\multicolumn{5}{l}{\emph{Zero-shot per field (mean over 5 tickers)}}\\
 & \textsc{type} & \textsc{price} & \textsc{size} & $\Delta t$ \\ \midrule
mixture (\Fone--\Fthree) & 1.53 & 1.30 & 1.60 & 3.07 \\
+ \Ffour{} enrichment & \textbf{1.48} & \textbf{1.29} & \textbf{0.91} & \textbf{2.81} \\
unigram (in-domain) & 1.50 & 1.59 & 0.90 & 1.95 \\ \midrule
\multicolumn{5}{l}{\emph{Fine-tune vs.\ scratch, mean over 3 held-out tickers}}\\
 & \multicolumn{1}{c}{5\% slice} & \multicolumn{1}{c}{20\%} &
   \multicolumn{1}{c}{80\%} & \multicolumn{1}{c}{zero-shot} \\ \midrule
fine-tuned & \textbf{1.140} & \textbf{1.092} & \textbf{1.062} & 1.585 \\
scratch & 1.275 & 1.188 & 1.115 & -- \\
advantage & $+$0.134 & $+$0.096 & $+$0.053 & \\
\bottomrule
\end{tabular}
\vspace*{-2mm}
\end{table}

\begin{figure}[t]
\centering
\includegraphics[width=\linewidth]{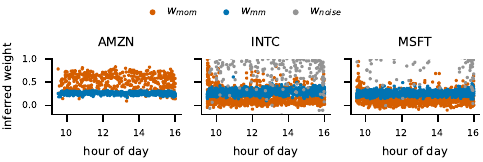}
\caption{Agent-composition probe read-outs on real NASDAQ streams. Each point estimates one agent weight on one 512-message window.}
\label{fig:realdata}
\vspace*{-2mm}
\end{figure}

As a qualitative illustration, Figure~\ref{fig:realdata} applies the
\Fthree{} agent-composition probe to the real streams: the inferred mix
varies systematically over the trading day. The read-out is an
interpretation of real flow through the simulator's lens, not a validated measurement, but it illustrates the kind of instrument the framework
provides.

\vspace{-1mm}

\section{Discussion and Conclusion}
\label{sec:discussion}

\textbf{Summary.} Next-token pretraining on order flow produces a
Bayes-optimal forecaster of market regimes, including under a
time-varying latent state (Section~\ref{sec:bayes}); the regime is
linearly decodable from hidden states beyond all baselines
(Section~\ref{sec:probes}); and both capabilities are created by regime
diversity, while capacity and token volume saturate early
(Section~\ref{sec:scaling}). On real data, transfer depends on the simulator prior's coverage of real flow
(Section~\ref{sec:transfer}).

\textbf{Implications for practice.} Four consequences follow for the
construction and evaluation of LOB foundation models. First, average
next-token loss is a poor measure of what a model has learned: it sits close to an entropy floor
that small models already reach, so differences between large models
will not appear in this metric even when they exist elsewhere. Model
comparison should instead examine regime coverage, behavior out of
distribution, and downstream read-outs. Second, when assembling training
data, the number of distinct market regimes matters more than the number of tokens, and our diversity ablation gives a concrete example in which the
same token budget spans the range from memorization to full
generalization. Third, a pretrained model is a calibration instrument at
essentially no cost: a linear read-out of frozen activations estimated
simulator parameters more accurately than both maximum likelihood and a
supervised task-specific regressor. Finally, decomposing predictive loss
by message field shows which distributional properties a simulator
lacks, making the framework useful for simulator development as well.

\textbf{Limitations.} (i) The simulators are simpler than real
markets: we establish what next-token pretraining can recover when the
generator is known, not that real order flow has an identifiable regime
of this kind. Real flow plausibly also has higher excess entropy, which
would extend the scaling regime beyond the saturation we observe. (ii) One structural rule, the cancellation floor,
breaks exact conditional independence in \Fone{}, with measured effect
below a millinat and reported where relevant. (iii) The scaling-law
exponents are poorly identified in our setting, a property of
known-floor regimes, not an estimate for real LOB data. (iv) The
real-data evaluation covers five tickers on one 2012 trading day, a
controlled analysis of transfer, not evidence about current markets.
(v) Closed-loop generation from the agent-based family shows a genuine
spread mismatch, so the realism of learned agent dynamics under feedback remains open. Our rollouts span 2{,}048 steps without exogenous order injection, and longer horizons with an injected agent (e.g., to
measure market impact) are future work. (vi) All results use one fixed tokenization. Bin edges were checked for balance, but sensitivity to the token schema
is not studied systematically.

\textbf{Future work.} Natural extensions include richer simulator
families, systematic study of tokenization choices, transfer experiments
at larger real-data scale, causal tests of whether the decoded regime
information is used (via residual-stream interventions), and comparisons against task-specific models
on downstream tasks, for which the framework provides exact optimal
policies as references.
\vspace{-1mm}

\balance
\bibliographystyle{ACM-Reference-Format}
\bibliography{refs}

\begin{thebibliography}{25}


\ifx \showCODEN    \undefined \def \showCODEN     #1{\unskip}     \fi
\ifx \showISBNx    \undefined \def \showISBNx     #1{\unskip}     \fi
\ifx \showISBNxiii \undefined \def \showISBNxiii  #1{\unskip}     \fi
\ifx \showISSN     \undefined \def \showISSN      #1{\unskip}     \fi
\ifx \showLCCN     \undefined \def \showLCCN      #1{\unskip}     \fi
\ifx \shownote     \undefined \def \shownote      #1{#1}          \fi
\ifx \showarticletitle \undefined \def \showarticletitle #1{#1}   \fi
\ifx \showURL      \undefined \def \showURL       {\relax}        \fi
\providecommand\bibfield[2]{#2}
\providecommand\bibinfo[2]{#2}
\providecommand\natexlab[1]{#1}
\providecommand\showeprint[2][]{arXiv:#2}

\bibitem[Aky{\"u}rek et~al\mbox{.}(2023)]%
        {akyurek2023what}
\bibfield{author}{\bibinfo{person}{Ekin Aky{\"u}rek}, \bibinfo{person}{Dale
  Schuurmans}, \bibinfo{person}{Jacob Andreas}, \bibinfo{person}{Tengyu Ma},
  {and} \bibinfo{person}{Denny Zhou}.} \bibinfo{year}{2023}\natexlab{}.
\newblock \showarticletitle{What Learning Algorithm Is In-Context Learning?
  {Investigations} with Linear Models}. In
  \bibinfo{booktitle}{\emph{International Conference on Learning
  Representations (ICLR)}}.
\newblock


\bibitem[Alain and Bengio(2016)]%
        {alain2016understanding}
\bibfield{author}{\bibinfo{person}{Guillaume Alain} {and}
  \bibinfo{person}{Yoshua Bengio}.} \bibinfo{year}{2016}\natexlab{}.
\newblock \showarticletitle{Understanding Intermediate Layers Using Linear
  Classifier Probes}.
\newblock \bibinfo{journal}{\emph{arXiv preprint arXiv:1610.01644}}
  (\bibinfo{year}{2016}).
\newblock


\bibitem[Bacry et~al\mbox{.}(2015)]%
        {bacry2015hawkes}
\bibfield{author}{\bibinfo{person}{Emmanuel Bacry}, \bibinfo{person}{Iacopo
  Mastromatteo}, {and} \bibinfo{person}{Jean-Fran{\c c}ois Muzy}.}
  \bibinfo{year}{2015}\natexlab{}.
\newblock \showarticletitle{Hawkes Processes in Finance}.
\newblock \bibinfo{journal}{\emph{Market Microstructure and Liquidity}}
  \bibinfo{volume}{1}, \bibinfo{number}{1} (\bibinfo{year}{2015}),
  \bibinfo{pages}{1550005}.
\newblock


\bibitem[Belinkov(2022)]%
        {belinkov2022probing}
\bibfield{author}{\bibinfo{person}{Yonatan Belinkov}.}
  \bibinfo{year}{2022}\natexlab{}.
\newblock \showarticletitle{Probing Classifiers: Promises, Shortcomings, and
  Advances}.
\newblock \bibinfo{journal}{\emph{Computational Linguistics}}
  \bibinfo{volume}{48}, \bibinfo{number}{1} (\bibinfo{year}{2022}),
  \bibinfo{pages}{207--219}.
\newblock


\bibitem[Byrd et~al\mbox{.}(2020)]%
        {byrd2020abides}
\bibfield{author}{\bibinfo{person}{David Byrd}, \bibinfo{person}{Maria
  Hybinette}, {and} \bibinfo{person}{Tucker~Hybinette Balch}.}
  \bibinfo{year}{2020}\natexlab{}.
\newblock \showarticletitle{{ABIDES}: Towards High-Fidelity Multi-Agent Market
  Simulation}.
\newblock \bibinfo{journal}{\emph{Proceedings of the 2020 ACM SIGSIM Conference
  on Principles of Advanced Discrete Simulation}} (\bibinfo{year}{2020}).
\newblock


\bibitem[Cont et~al\mbox{.}(2014)]%
        {cont2014price}
\bibfield{author}{\bibinfo{person}{Rama Cont}, \bibinfo{person}{Arseniy
  Kukanov}, {and} \bibinfo{person}{Sasha Stoikov}.}
  \bibinfo{year}{2014}\natexlab{}.
\newblock \showarticletitle{The Price Impact of Order Book Events}.
\newblock \bibinfo{journal}{\emph{Journal of Financial Econometrics}}
  \bibinfo{volume}{12}, \bibinfo{number}{1} (\bibinfo{year}{2014}),
  \bibinfo{pages}{47--88}.
\newblock


\bibitem[Garg et~al\mbox{.}(2022)]%
        {garg2022what}
\bibfield{author}{\bibinfo{person}{Shivam Garg}, \bibinfo{person}{Dimitris
  Tsipras}, \bibinfo{person}{Percy Liang}, {and} \bibinfo{person}{Gregory
  Valiant}.} \bibinfo{year}{2022}\natexlab{}.
\newblock \showarticletitle{What Can Transformers Learn In-Context? {A} Case
  Study of Simple Function Classes}. In \bibinfo{booktitle}{\emph{Advances in
  Neural Information Processing Systems}}.
\newblock


\bibitem[Hoffmann et~al\mbox{.}(2022)]%
        {hoffmann2022chinchilla}
\bibfield{author}{\bibinfo{person}{Jordan Hoffmann}, \bibinfo{person}{Sebastian
  Borgeaud}, \bibinfo{person}{Arthur Mensch}, {et~al\mbox{.}}}
  \bibinfo{year}{2022}\natexlab{}.
\newblock \showarticletitle{Training Compute-Optimal Large Language Models}. In
  \bibinfo{booktitle}{\emph{Advances in Neural Information Processing
  Systems}}.
\newblock


\bibitem[Hollmann et~al\mbox{.}(2023)]%
        {hollmann2023tabpfn}
\bibfield{author}{\bibinfo{person}{Noah Hollmann}, \bibinfo{person}{Samuel
  M{\"u}ller}, \bibinfo{person}{Katharina Eggensperger}, {and}
  \bibinfo{person}{Frank Hutter}.} \bibinfo{year}{2023}\natexlab{}.
\newblock \showarticletitle{{TabPFN}: A Transformer That Solves Small Tabular
  Classification Problems in a Second}. In
  \bibinfo{booktitle}{\emph{International Conference on Learning
  Representations (ICLR)}}.
\newblock


\bibitem[Huang and Polak(2011)]%
        {huang2011lobster}
\bibfield{author}{\bibinfo{person}{Ruihong Huang} {and} \bibinfo{person}{Tomas
  Polak}.} \bibinfo{year}{2011}\natexlab{}.
\newblock \showarticletitle{{LOBSTER}: Limit Order Book Reconstruction System}.
\newblock \bibinfo{journal}{\emph{SSRN 1977207}} (\bibinfo{year}{2011}).
\newblock


\bibitem[Kaplan et~al\mbox{.}(2020)]%
        {kaplan2020scaling}
\bibfield{author}{\bibinfo{person}{Jared Kaplan}, \bibinfo{person}{Sam
  McCandlish}, \bibinfo{person}{Tom Henighan}, {et~al\mbox{.}}}
  \bibinfo{year}{2020}\natexlab{}.
\newblock \showarticletitle{Scaling Laws for Neural Language Models}.
\newblock \bibinfo{journal}{\emph{arXiv preprint arXiv:2001.08361}}
  (\bibinfo{year}{2020}).
\newblock


\bibitem[Kawawa-Beaudan et~al\mbox{.}(2026)]%
        {tradefm2026}
\bibfield{author}{\bibinfo{person}{Maxime Kawawa-Beaudan},
  \bibinfo{person}{Srijan Sood}, \bibinfo{person}{Kassiani Papasotiriou},
  \bibinfo{person}{Daniel Borrajo}, {and} \bibinfo{person}{Manuela Veloso}.}
  \bibinfo{year}{2026}\natexlab{}.
\newblock \showarticletitle{{TradeFM}: A Generative Foundation Model for
  Trade-flow and Market Microstructure}.
\newblock \bibinfo{journal}{\emph{arXiv preprint arXiv:2602.23784}}
  (\bibinfo{year}{2026}).
\newblock


\bibitem[Li et~al\mbox{.}(2025)]%
        {mars2025}
\bibfield{author}{\bibinfo{person}{Junjie Li}, \bibinfo{person}{Yang Liu},
  \bibinfo{person}{Weiqing Liu}, \bibinfo{person}{Shikai Fang},
  \bibinfo{person}{Lewen Wang}, \bibinfo{person}{Chang Xu}, {and}
  \bibinfo{person}{Jiang Bian}.} \bibinfo{year}{2025}\natexlab{}.
\newblock \showarticletitle{{MarS}: A Financial Market Simulation Engine
  Powered by Generative Foundation Model}. In
  \bibinfo{booktitle}{\emph{International Conference on Learning
  Representations (ICLR)}}.
\newblock


\bibitem[Mikulik et~al\mbox{.}(2020)]%
        {mikulik2020meta}
\bibfield{author}{\bibinfo{person}{Vladimir Mikulik},
  \bibinfo{person}{Gr{\'e}goire Del{\'e}tang}, \bibinfo{person}{Tom McGrath},
  \bibinfo{person}{Tim Genewein}, \bibinfo{person}{Miljan Martic},
  \bibinfo{person}{Shane Legg}, {and} \bibinfo{person}{Pedro~A. Ortega}.}
  \bibinfo{year}{2020}\natexlab{}.
\newblock \showarticletitle{Meta-trained Agents Implement {Bayes}-optimal
  Agents}. In \bibinfo{booktitle}{\emph{Advances in Neural Information
  Processing Systems}}.
\newblock


\bibitem[M{\"u}ller et~al\mbox{.}(2022)]%
        {muller2022transformers}
\bibfield{author}{\bibinfo{person}{Samuel M{\"u}ller}, \bibinfo{person}{Noah
  Hollmann}, \bibinfo{person}{Sebastian Pineda~Arango}, \bibinfo{person}{Josif
  Grabocka}, {and} \bibinfo{person}{Frank Hutter}.}
  \bibinfo{year}{2022}\natexlab{}.
\newblock \showarticletitle{Transformers Can Do {Bayesian} Inference}. In
  \bibinfo{booktitle}{\emph{International Conference on Learning
  Representations (ICLR)}}.
\newblock


\bibitem[Nagy et~al\mbox{.}(2025)]%
        {lobbench2025}
\bibfield{author}{\bibinfo{person}{Peer Nagy}, \bibinfo{person}{Sascha Frey},
  \bibinfo{person}{Kang Li}, \bibinfo{person}{Bidipta Sarkar},
  \bibinfo{person}{Svitlana Vyetrenko}, \bibinfo{person}{Stefan Zohren},
  \bibinfo{person}{Anisoara Calinescu}, {and} \bibinfo{person}{Jakob
  Foerster}.} \bibinfo{year}{2025}\natexlab{}.
\newblock \showarticletitle{{LOB-Bench}: Benchmarking Generative {AI} for
  Finance -- an Application to Limit Order Book Data}. In
  \bibinfo{booktitle}{\emph{International Conference on Machine Learning
  (ICML)}}.
\newblock


\bibitem[Nagy et~al\mbox{.}(2023)]%
        {nagy2023generative}
\bibfield{author}{\bibinfo{person}{Peer Nagy}, \bibinfo{person}{Sascha Frey},
  \bibinfo{person}{Silvia Sapora}, {et~al\mbox{.}}}
  \bibinfo{year}{2023}\natexlab{}.
\newblock \showarticletitle{Generative {AI} for End-to-End Limit Order Book
  Modelling: A Token-Level Autoregressive Generative Model of Message Flow
  Using a Deep State Space Network}. In \bibinfo{booktitle}{\emph{Proceedings
  of the 4th ACM International Conference on AI in Finance (ICAIF)}}.
\newblock


\bibitem[Ogata(1981)]%
        {ogata1981lewis}
\bibfield{author}{\bibinfo{person}{Yosihiko Ogata}.}
  \bibinfo{year}{1981}\natexlab{}.
\newblock \showarticletitle{On {Lewis}' Simulation Method for Point Processes}.
\newblock \bibinfo{journal}{\emph{IEEE Transactions on Information Theory}}
  \bibinfo{volume}{27}, \bibinfo{number}{1} (\bibinfo{year}{1981}),
  \bibinfo{pages}{23--31}.
\newblock


\bibitem[Ortega et~al\mbox{.}(2019)]%
        {ortega2019meta}
\bibfield{author}{\bibinfo{person}{Pedro~A. Ortega}, \bibinfo{person}{Jane~X.
  Wang}, \bibinfo{person}{Mark Rowland}, \bibinfo{person}{Tim Genewein},
  {et~al\mbox{.}}} \bibinfo{year}{2019}\natexlab{}.
\newblock \showarticletitle{Meta-learning of Sequential Strategies}.
\newblock \bibinfo{journal}{\emph{arXiv preprint arXiv:1905.03030}}
  (\bibinfo{year}{2019}).
\newblock


\bibitem[Ravent{\'o}s et~al\mbox{.}(2023)]%
        {raventos2023pretraining}
\bibfield{author}{\bibinfo{person}{Allan Ravent{\'o}s},
  \bibinfo{person}{Mansheej Paul}, \bibinfo{person}{Feng Chen}, {and}
  \bibinfo{person}{Surya Ganguli}.} \bibinfo{year}{2023}\natexlab{}.
\newblock \showarticletitle{Pretraining Task Diversity and the Emergence of
  Non-{Bayesian} In-Context Learning for Regression}. In
  \bibinfo{booktitle}{\emph{Advances in Neural Information Processing
  Systems}}.
\newblock


\bibitem[Sirignano and Cont(2019)]%
        {sirignano2021universal}
\bibfield{author}{\bibinfo{person}{Justin Sirignano} {and}
  \bibinfo{person}{Rama Cont}.} \bibinfo{year}{2019}\natexlab{}.
\newblock \showarticletitle{Universal Features of Price Formation in Financial
  Markets: Perspectives from Deep Learning}.
\newblock \bibinfo{journal}{\emph{Quantitative Finance}} \bibinfo{volume}{19},
  \bibinfo{number}{9} (\bibinfo{year}{2019}), \bibinfo{pages}{1449--1459}.
\newblock


\bibitem[Smith et~al\mbox{.}(2003)]%
        {smith2003statistical}
\bibfield{author}{\bibinfo{person}{Eric Smith}, \bibinfo{person}{J.~Doyne
  Farmer}, \bibinfo{person}{L{\'a}szl{\'o} Gillemot}, {and}
  \bibinfo{person}{Supriya Krishnamurthy}.} \bibinfo{year}{2003}\natexlab{}.
\newblock \showarticletitle{Statistical Theory of the Continuous Double
  Auction}.
\newblock \bibinfo{journal}{\emph{Quantitative Finance}} \bibinfo{volume}{3},
  \bibinfo{number}{6} (\bibinfo{year}{2003}), \bibinfo{pages}{481--514}.
\newblock


\bibitem[von Oswald et~al\mbox{.}(2023)]%
        {vonoswald2023transformers}
\bibfield{author}{\bibinfo{person}{Johannes von Oswald},
  \bibinfo{person}{Eyvind Niklasson}, \bibinfo{person}{Ettore Randazzo},
  \bibinfo{person}{Jo{\~a}o Sacramento}, \bibinfo{person}{Alexander
  Mordvintsev}, \bibinfo{person}{Andrey Zhmoginov}, {and} \bibinfo{person}{Max
  Vladymyrov}.} \bibinfo{year}{2023}\natexlab{}.
\newblock \showarticletitle{Transformers Learn In-Context by Gradient Descent}.
  In \bibinfo{booktitle}{\emph{International Conference on Machine Learning
  (ICML)}}.
\newblock


\bibitem[Xie et~al\mbox{.}(2022)]%
        {xie2021explanation}
\bibfield{author}{\bibinfo{person}{Sang~Michael Xie}, \bibinfo{person}{Aditi
  Raghunathan}, \bibinfo{person}{Percy Liang}, {and} \bibinfo{person}{Tengyu
  Ma}.} \bibinfo{year}{2022}\natexlab{}.
\newblock \showarticletitle{An Explanation of In-context Learning as Implicit
  {Bayesian} Inference}. In \bibinfo{booktitle}{\emph{International Conference
  on Learning Representations (ICLR)}}.
\newblock


\bibitem[Zhang et~al\mbox{.}(2019)]%
        {zhang2019deeplob}
\bibfield{author}{\bibinfo{person}{Zihao Zhang}, \bibinfo{person}{Stefan
  Zohren}, {and} \bibinfo{person}{Stephen Roberts}.}
  \bibinfo{year}{2019}\natexlab{}.
\newblock \showarticletitle{{DeepLOB}: Deep Convolutional Neural Networks for
  Limit Order Books}.
\newblock \bibinfo{journal}{\emph{IEEE Transactions on Signal Processing}}
  \bibinfo{volume}{67}, \bibinfo{number}{11} (\bibinfo{year}{2019}),
  \bibinfo{pages}{3001--3012}.
\newblock


\end{thebibliography}

\end{document}